\documentclass[11pt,a4paper]{article}
\usepackage[hyperref]{acl2017}
\usepackage{times}
\usepackage{latexsym}
\usepackage{subcaption}
\usepackage{url}
\usepackage{graphicx}
\usepackage{flushend}
\usepackage{balance}

\usepackage{epsfig}
\usepackage{amsmath}
\usepackage{amssymb}
\usepackage{booktabs}
\usepackage{tikz}
\usepackage{gensymb}
\usepackage{balance}
\usepackage{siunitx}

\usepackage{etoolbox}
\makeatletter
\newcount\c@additionalboxlevel
\newcount\c@maxboxlevel
\patchcmd\@combinedblfloats{\box\@outputbox}{%
	\stepcounter{additionalboxlevel}%
	\box\@outputbox
}{}{\errmessage{\noexpand\@combinedblfloats could not be patched}}

\AtBeginShipout{%
	\ifnum\value{additionalboxlevel}>\value{maxboxlevel}%
	\typeout{Warning: maxboxlevel might be too small, increase to %
		\the\value{additionalboxlevel}%
	}%
	\fi 
	\@whilenum\value{additionalboxlevel}<\value{maxboxlevel}\do{%
		\typeout{* Additional boxing of page `\thepage'}%
		\setbox\AtBeginShipoutBox=\hbox{\copy\AtBeginShipoutBox}%
		\stepcounter{additionalboxlevel}%
	}%
	\setcounter{additionalboxlevel}{0}%
}
\makeatother

\aclfinalcopy 

\def\confidential{}

\def\aclruler#1{\makevruler[14.17pt][#1][1][3][\textheight]\usebox{\aclrulerbox}}
\def\leftoffset{-2.1cm} 
\def\rightoffset{17.5cm} 
\AddToShipoutPicture{%
\aclruleroffset=\textheight
\advance\aclruleroffset4pt
  \AtTextUpperLeft{
    \put(0,\LenToUnit{1cm}){\parbox{\textwidth}{\centering\naaclhv\confidential}}
  }
}

\title{End-to-End Spoken Language Translation}

\author{Michelle Guo \qquad \, \,  Albert Haque \qquad Prateek Verma \\
  Technical Report\thanks{\; This technical report was originally done in 2017 as a course project and can be found online at Stanford University \href{http://web.stanford.edu/class/cs224s/reports/Michelle_Guo.pdf}{here}. It was uploaded to arXiv in 2019 for archival purposes.} \\
  Stanford University (2017)}

\begin{document}
\maketitle
\begin{abstract}
    In this paper, we address the task of spoken language understanding.
	We present a method for translating spoken sentences from one language into spoken sentences in another language.
	Given spectrogram-spectrogram pairs, our model can be trained completely from scratch to translate unseen sentences.
	Our method consists of a pyramidal-bidirectional recurrent network combined with a convolutional network to output sentence-level spectrograms in the target language.
	Empirically, our model achieves competitive performance with state-of-the-art methods on multiple languages and can generalize to unseen speakers.
\end{abstract}

\section{Introduction}

Humans are able to seamlessly process different language representations despite syntactic, acoustic, and semantic variations.
Recent studies from the neurolinguistics literature suggest that humans have distinct submodules for different linguistic functions \cite{hickok2007cortical}.
One could suspect word-level and sentence-level processing may occur in entirely different cortical regions \cite{embick2000syntactic}.
One theory for acoustic-language invariance posits that different languages do activate different submodules but instead of being linked to proficiency, these submodules are linked to general language processing \cite{newman2012influence}.

\begin{figure}[t]
	\centering
	\includegraphics[width=1.0\linewidth]{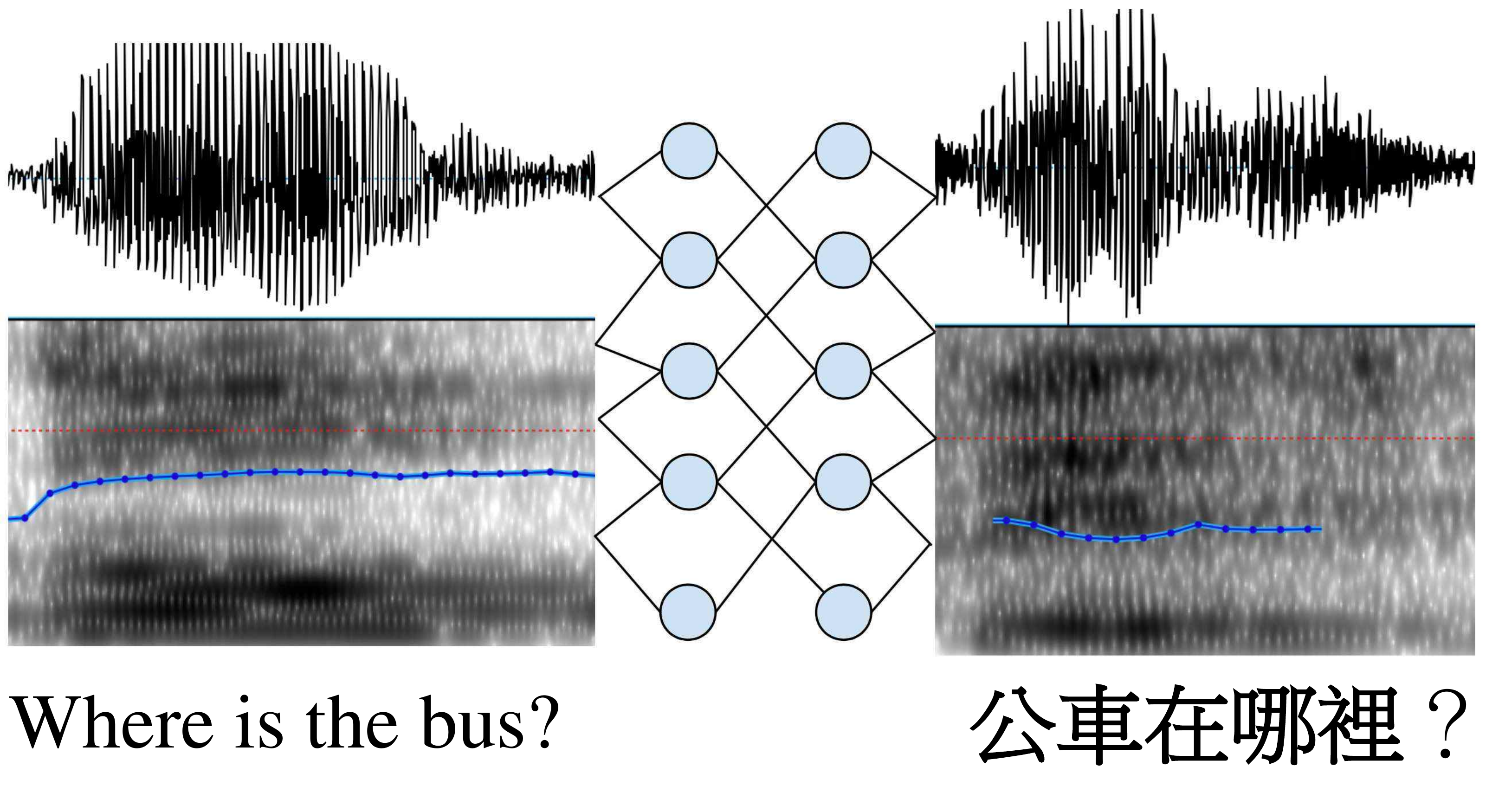}
	\caption{Our goal is to translate spoken sentences from one language to another, end-to-end.}
	\label{fig:pull}
\end{figure}

Inspired by humans, modern machine translation systems often use a word-level model to aid in the translation process \cite{luong2015effective, zou2013bilingual}.
In the case of text-based translation, learned word vectors or one-hot embeddings are the primary means of representing natural language \cite{pennington2014glove, mikolov2013efficient}.
For speech and acoustic inputs however, word or phone embeddings are often used as a training convenience to provide multiple sources of information gradient flow to the model \cite{bourlard1990continuous, wilpon1990automatic}.
Spectrograms remain the dominant acoustic representation for both phoneme and word-level tasks since the high sampling rate and dimensionality of waveforms is difficult to model \cite{van2016wavenet}.

Intuitively, sentence-level and phrase-level representations seem to be a more powerful modeling tool for some tasks \cite{wilson2005recognizing}.
However, the work on sentence- and phrase-level features in the context of automatic speech recognition and machine translation has been limited.
The primary challenge is the ambiguities and combinatorics associated with long-term temporal dependencies, especially for longer sentences containing ten or twenty words in each language \cite{xu2015show, venugopalan2014translating}.
Another challenge is limited data: there are no publicly available spoken language translation datasets that contain multiple speakers and multiple languages.

In this paper, we address the task of spoken language understanding, distinct from speech recognition (speech-to-text) and speech synthesis (text-to-speech).
Specifically, we propose a method for end-to-end\footnote{Although the output of our model is not a waveform, we follow the definitions from \citet{wang2017tacotron}.} spoken language translation.
The input to our model is an acoustic sentence in one language and the output is an acoustic representation of the same sentence translated into a different language.
Given the importance of the temporal speech cortex in the human brain \cite{geschwind1968human}, the foundation of our model is a recurrent network.
Our method operates on spectrograms and combines the Listen-Attend-Spell model  \cite{chan2016listen} with the TacoTron \cite{wang2017tacotron} architecture.

Our contributions are as follows. First, we propose a fully differentiable sentence-level acoustic model for translating sentences across languages.
Empirically, we show the success of our model through various ablation studies and subjective evaluation tests.
Second, we collect and publicly release a dataset containing twelve speakers uttering words from Spanish, Mandarin, Hindi, and English.
Our dataset collection technique allows for efficient audio acquisition and programmatic generation of grammatically correct spoken sentences.
This is an efficient and scalable way to collect large datasets for our spoken language translation task.

\section{Related Work}

\textbf{Speech-to-Text}. Also known as speech recognition, the task of speech-to-text is to accept as input an audio signal and output a text-based representation of the words spoken in the input.
Recent work in speech recognition often employs Hidden Markov Models (HMMs) or deep networks.

HMM-based methods \cite{bahl1986maximum} attempt to model sequences of acoustic features extracted from audio input.
One of the simplest features are mel-frequency cepstral coefficients (MFCCs)  \cite{logan2000mel} and attempt to model the responses in the human ear.
Using the independence assumption, HMMs model words by analyzing a small temporal context window of acoustic features \cite{levinson1986continuously}.
Hidden Markov models have been successful not only in speech recognition but also emotion recognition \cite{schuller2003hidden, nwe2003speech}, gender classification \cite{konig1992gdnn}, and accent classification \cite{arslan1996language}.

Deep learning methods generally make use of recurrent networks.
One such example is the Connectionist Temporal Classification for speech recognition \cite{graves2006connectionist, graves2014towards}. More recently, attention-based models \cite{xu2015show} such as Listen, Attend, and Spell \cite{chan2016listen} proposed a system using a sequence-to-sequence model with attention.

\textbf{Text-to-Speech.} Also known as speech synthesis, text-to-speech (TTS) systems have just recently started to show promising results.
It has been shown that a pre-trained HMM combined with a sequence-to-sequence model can learn appropriate alignments \cite{wang2016first}.
Unfortunately this is not end-to-end as it predicts vocoder parameters and it is unclear how much performance is gained from the HMM aligner.
Char2wav \cite{sotelo2017char2wav} is another method, trained end-to-end on character inputs to produce audio but also predicts vocoder parameters.

DeepVoice \cite{arik2017deep} improves on this by replacing nearly all components in a standard TTS pipeline with neural networks. While this is closer towards a fully differentiable solution, each component is trained in isolation with different optimization objectives.
WaveNet \cite{van2016wavenet} is a powerful generative model of audio and generates realistic output speech.
However it is slow due to audio sample-level autoregressive nature and requires domain-specific linguistic feature engineering.

\begin{figure*}[t]
    \begin{center}
    \vspace{-5mm}
    \includegraphics[width=0.8\linewidth]{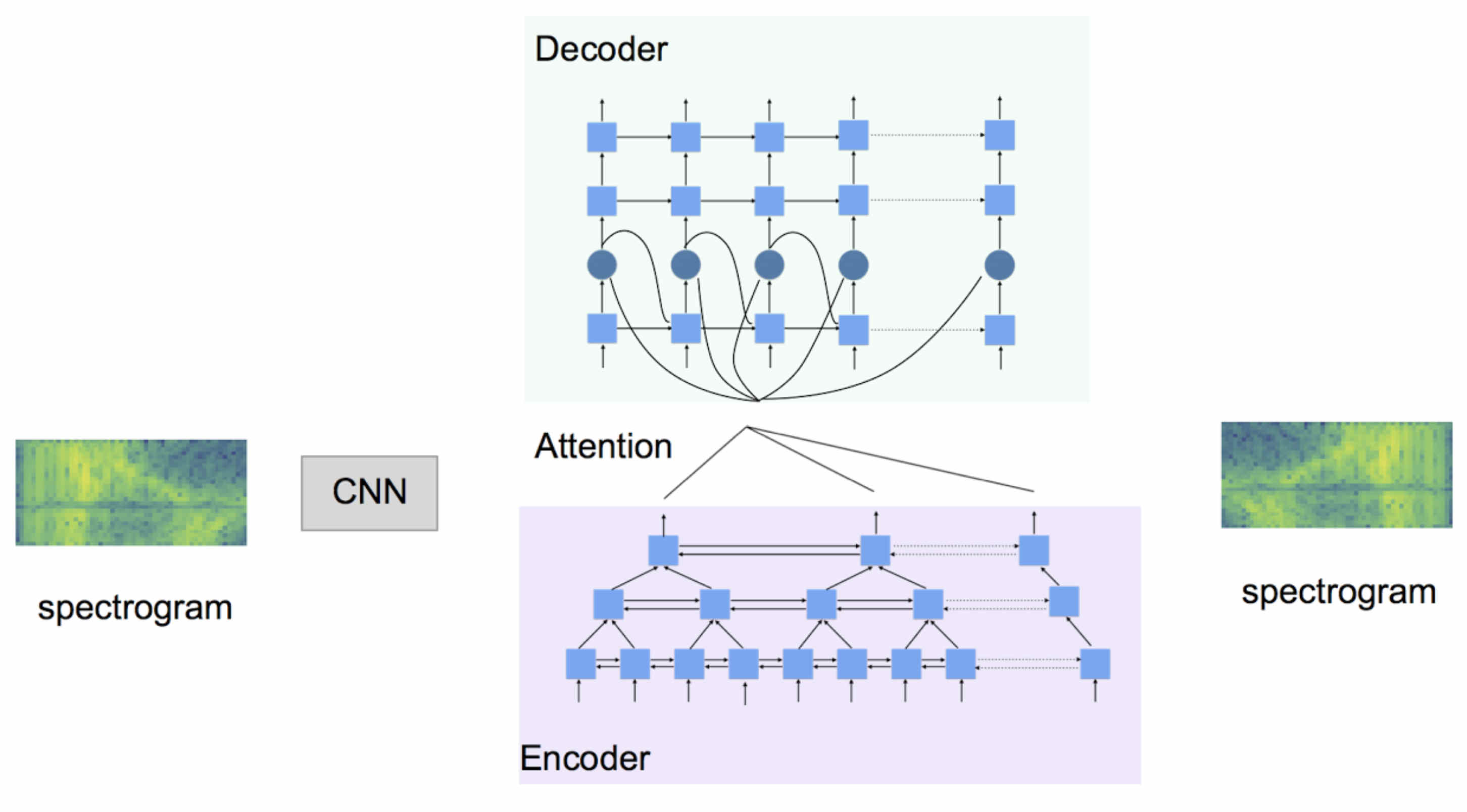}
    \end{center}
    \caption{Overview of our model. The inputs and outputs of our model are spectrograms. We use a convolutional network to encode temporal context windows in the input. Our encoder consists of a pyramidal bidirectional RNN which captures acoustic features and multiple levels of temporal resolution. The decoder is equipped with attention and outputs multiple spectrogram slices at a time.}
    \label{fig:method}
\end{figure*}

Most similar to our work is Tacotron \cite{wang2017tacotron}.
In this work, the authors move even closer to a fully differentiable system.
The input to Tacotron is a sequence of character embeddings and the output is a linear-scale spectrogram.
After applying Griffin-Lim phase reconstruction \cite{nawab1983signal}, the waveform is generated.

\textbf{Text-Based Machine Translation.} State-of-the-art machine translation methods generally come from a family of encoder-decoder, or sequence-to-sequence, models where the input is encoded into a fixed-length representation and is then decoded into the target sequence \cite{cho2014learning, sutskever2014sequence}.
In \cite{cho2014learning}, a recurrent network was used as an encoder-decoder model for machine translation.
While the presented results were positive, Bahdanau et al. argued that a fixed-length representation is a bottleneck during learning \cite{bahdanau2014neural}. They proposed the use of an attention mechanism \cite{luong2015effective, xu2015show} to improve performance.
Following this trend of attention-based sequence-to-sequence models, we now propose a model for end-to-end spoken language translation.

\section{Method}

Our goal is end-to-end spoken language translation.
Given an input spectrogram of a sentence spoken in one language, our model outputs a spectrogram of the same sentence spoken in a different language.
The backbone of our model is a sequence-to-sequence architecture with attention \cite{bahdanau2014neural, vinyals2015grammar}.
Figure \ref{fig:method} shows our model which consists of a convolutional network fuse temporal context, a pyramidal bidirectional encoder to capture acoustic features at multiple levels of temporal granularity, and a decoder with attention.
The model is fully differntiable and can be trained with modern optimization methods.

\subsection{Encoder}
\textbf{Convolutional Network.} The first step of our model is to learn an appropriate representation of the spectrogram input.
In its original form, the input consists of on the order of hundreds of timesteps. For longer sentences, this can easily expand to the order of thousands.
Each timestep is associated with a feature vector, also containing hundreds of real values.

Modeling the full spectrogram would require unrolling of the encoder RNN for an infeasibily large number of timesteps \cite{sainath2015learning}.
Even with truncated backpropagation through time \cite{haykin2001kalman}, this would be a challenging task on large datasets.
Inspired by the Convolutional, Long Short-Term Memory Deep Neural Network (CLDNN) \cite{sainath2015learning} approach, we use a convolutional network to reduce the temporal length of the input by using a learned convolutional filter bank.
The stride, or hop size, controls the degree of length reduction.

\textbf{Pyramidal Bidirectional Recurrent Network.} 
Many successful uses of RNNs have been on sentence-level outputs such as  machine translation \cite{bahdanau2014neural} and image captioning \cite{karpathy2015deep}.
The outputs are generally on the order of tens of timesteps.
Given the filter activations from our convolutional network, the input is of a smaller dimensionality and can be fed into our encoder network, however, it can still have hundreds of timesteps.
In Listen, Attend, and Spell (LAS) \cite{chan2016listen}, it was shown that a multi-layer RNN struggles to extract appropriate representations from the input due to the large number of timesteps.

Inspired by the Clockwork RNN \cite{koutnik2014clockwork}, we use a pyramidal RNN to address the issue of learning from a large number of timesteps \cite{chan2016listen}.
A pyramidal RNN is the same as a standard multi-layer RNN but instead of each layer simply accepting the input from the previous layer, successively higher layers in the network only compute, or ``tick," during particular timesteps.
This allows different layers of the RNN to operate at different temporal scales. Formally, let $h_i^j$ denote the hidden state of a bidirectional LSTM (BLSTM) at the $i$-th timestep of the $j$-th layer:
\begin{equation}\label{eq:blstm}
    h_i^j = \textrm{BLSTM}(h_{i-1}^j, h_i^{j-1})
\end{equation}
For a pyramidal BLSTM, the outputs from the lower layers, which contain high-resolution temporal information, are concatenated:
\begin{equation}\label{eq:pblstm}
    h_i^j = \textrm{pBLSTM}(h_{i-1}^j, \left[ h_{2i}^{j-1}, h_{2i+1}^{j-1} \right])
\end{equation}
In (\ref{eq:pblstm}), the output of a pBLSTM unit is now a function of not only its previous hidden state, but also the outputs from two timesteps from the layer below.
In LAS and our method, we reduce the time resolution by a factor of two for each layer.

Not only does the pyramidal RNN provide higher-level temporal features, but it also reduces the inference complexity.
Only the first layer processes each input timestep as opposed to all layers.

\subsection{Decoder}
\textbf{Attention}. Learning long-range temporal dependencies can be challenging \cite{bengio1994learning}.
To aid this process, we use an attention-based LSTM transducer \cite{chorowski2015attention}. Specifically, we selected a tanh attention mechanism \cite{bahdanau2014neural}.

At each timestep, the transducer produces a probability distribution over the next character conditioned on all the previously seen inputs.
The distribution for $y_i$ is a function of the decoder state $s_i$ and context $c_i$.
The decoder state $s_i$ is a function of the previous state $s_{i−1}$, the previously emitted character $y_{i−1}$ and context $c_{i−1}$. The context vector $c_i$ is produced by an attention mechanism \cite{chan2016listen}. Specifically, we define:
\begin{equation}
    \textrm{Context Vector } = c_i = \sum\limits_u \alpha_{i,u} h_u
\end{equation}
where attention is defined as the alignment between the current decoder frame $i$ and a frame $j$ from the encoder input:
\begin{equation}
    \alpha_{i,j} = \frac{exp(e_{i,j})}{\sum_{j=1}^{L}exp(e_{i,j}}
\end{equation}
and where the score between the output of the encoder or the hidden states, $h_j$, and the previous state of the decoder cell, $s_{i-1}$ is computed with: $e_{i,u} = \langle \phi(s_i), \varphi(h_u)\rangle$ where $\phi$ and $\varphi$ are sub-networks, e.g. multi-layer perceptrons:
\begin{equation}
    e_{i,j} = w^T \tanh(Ws_{i-1} + Vh_j + b)
\end{equation}
for with learnable parameters $w$, $W$ and $V$. 

\textbf{Multiple Output Prediction}. The decoder faces similar timestep challenges as the input. Because we are predicting entire sentences, the number of output timesteps is on the order of hundreds to thousands.
We follow a method proposed by Tacotron \cite{wang2017tacotron} to remedy this issue.
Specifically, at each decoder timestep, we predict $r$ spectrogram slices. This reduces the number of output timesteps by a factor of $r$.
Other benefits include reduced number of parameters (see Table \ref{table:ablation}), faster training time, and better spectrogram reconstruction performance.

\textbf{Spectrogram to Waveform}. Our model predicts spectrogram magnitudes only. To produce a waveform, we need both the magnitude and the phase components. Since our model does not predict phase, we use our predicted magnitude and apply a Griffin-Lim phase recovery \cite{nawab1983signal} to generate the final waveform of the translated sentence.

\subsection{Optimization}

Due to the highly complex nature of our method, we assessed the performance of multiple loss objectives in isolation and in parallel.
Because the output of our model is a spectrogram, we use the standard L2 loss:
\begin{equation}
    \mathcal{L}_{\ell 2}(Y, \hat{Y}) = \frac{1}{n} \sum\limits_{i=1}^{n} || Y^{(i)} - \hat{Y}^{(i)} ||
\end{equation}
where $Y^{(i)}$ denotes the ground truth spectrogram for training example $i$ and $\hat{Y}^{(i)}$ denotes the predicted spectrogram for example $i$.

\begin{figure*}[t]
	\centering
	\includegraphics[width=0.8\linewidth]{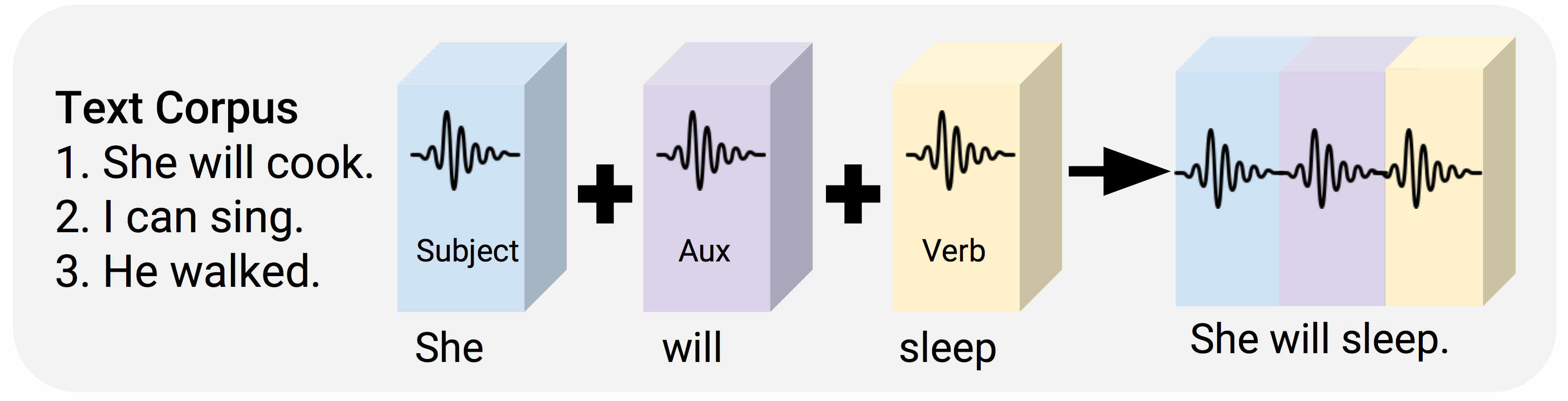}
	\caption{Concatenative sentence generation. Recordings were collected of multiple people speaking individual words in different languages. Using fixed vocabulary size, we selected real-world text sentences from public text corpora. We then programatically constructed full spoken sentences by concatenating the individual recorded words.}
	\label{fig:concat}
\end{figure*}

We also experimented with the Kullback-Leibler divergence as our optimization criterion:
\begin{equation}
    \mathcal{L}_{\textrm{KL}}(P, Q) = \sum\limits_{i} P(i) \log \frac{P(i)}{Q(i)}
\end{equation}
where $P$ and $Q$ denote the predicted and ground truth output distributions, respectively.

In cases where the dimensionality is high, direct regression may be unable to efficiently learn \cite{van2016wavenet}.
One solution is to quantize the output and treat the task as a classification problem \cite{oehler1995combining}.
Instead of predicting real-valued outputs corresponding to the output spectrogram, we quantize the ground truth label into $K=256$ bins and optimze the model as a classification task.
\begin{equation}
    \mathcal{L}_{\textrm{xent}}(P, Q) = -\sum\limits_x P(x) \log Q(x)
\end{equation}
where $P$ and $Q$ denote the predicted and ground truth output distributions, respectively.
An added benefit is that we can push down the probability of incorrect bins while simultaneously increasing the probability of the correct output.

The KL-divergence loss, L2 loss, and cross-entropy loss were weighted according to mixing coefficients $\lambda_1$, $\lambda_2$, and $\lambda_3$.

\section{Experiments}

Our goal is end-to-end spoken language translation.
In general, end-to-end models require large amounts of training data.
Because we collected our own dataset, we must efficiently collect a large number of training examples.
This was done by concatenatively generating sentences from real-world recordings of individual words.

\subsection{Pearl Dataset}

We collect a new spoken language translation dataset, titled \textit{Pearl}.
The dataset consists of grammatically correct input and output sentences spoken by the same speaker.
The input is spoken in a different language from the output.
A total of twelve speakers (8 male, 4 female) contributed to the dataset.
The languages include Hindi, Mandarin, English, and Spanish.

\textbf{Vocabulary}. First, a pre-determined vocabulary of 100 English words was determined such that the vocabulary list contains words from multiple grammatical categories (e.g., noun, verb).
Each speaker was instructed to speak one word at a time from the vocabulary list.
This can be repeated multiple times per word to collect diverse pitch tracks and intonation.

\begin{figure*}[t]
	\begin{center}
		\includegraphics[width=1\linewidth]{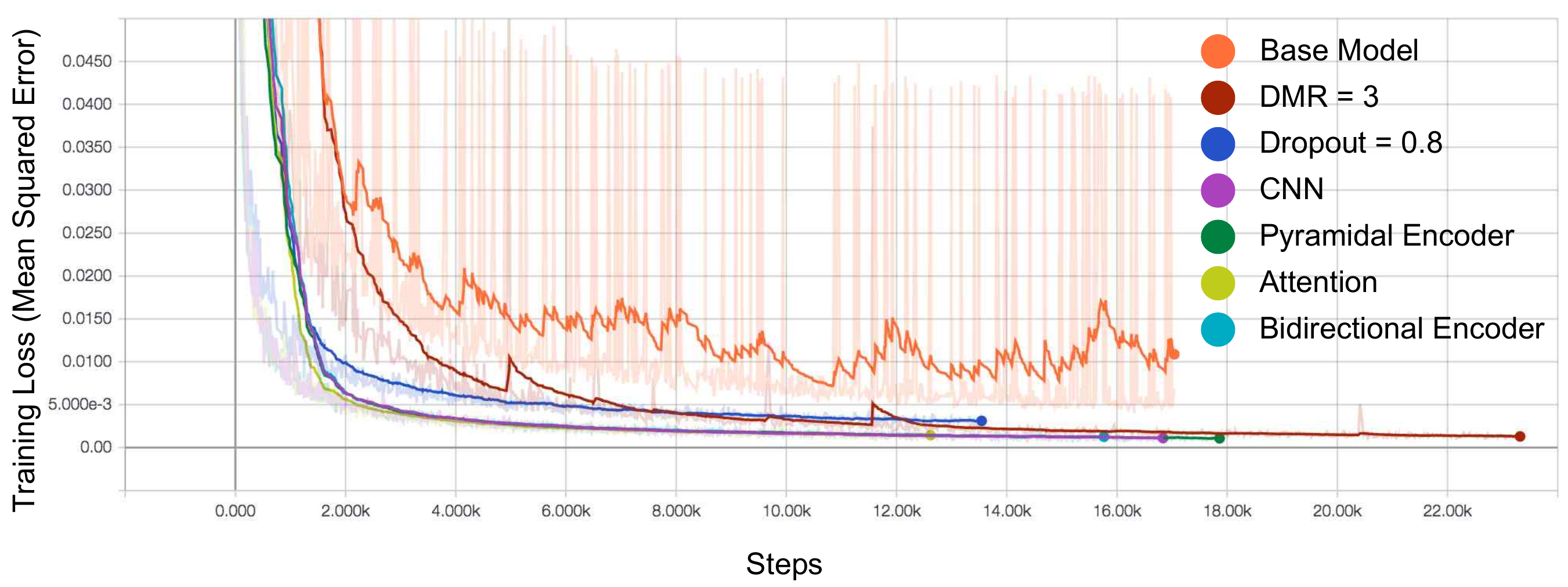}
	\end{center}
	\caption{Ablation Study of Model Components (ZH to EN). Various methods show an improvement in the reconstruction error on top of the base model. The above plot serves as a powerful debugging tool for analyzing each component of our model. The base model fluctuates during training and maintains the highest error.
	}
	\label{fig:plot}
\end{figure*}

\textbf{Concatenative Generation.}
First, sentences were constructed with a concatenative model using real-world sentences from the Facebook bAbI project \cite{li2016dialogue}. We constructed single-word inputs, bigrams, trigrames, and sentences consisting of five to ten words. Additionally, we extracted single words and single phones from the TIMIT dataset \cite{garofolo1993darpa}. These words were similarly concatenated to construct English sentences

Second, the vocabulary list was translated to other languages using the Google Translate API.
Each speaker was asked to speak words from a pre-determined vocabulary list consisting of 100 English words.
The English words were translated into other languages using the Google Translate API and were manually verified by our native speakers. Figure \ref{fig:concat} shows an overview of this process.

\subsection{Comparison with State-of-the-Art}
We performed ablation studies to assess the effect of the various components of our model when equipped with state-of-the-art modules.
Starting with a base encoder-decoder model, we introduce a single method.
The results of each model is shown in Figure \ref{table:ablation}.
Because we individually evaluated the performance of each state-of-the-art method in isolation, we can better understand our model's performance improvements.
We show the changes in parameter size as well as the model's spectrogram reconstruction error using mean-squared error across the different methods.

\begin{table}[t]
	\centering
	\begin{tabular}{l|ccc|c|cccccccc} \toprule
		Method & \# Parameters & Error\\ \midrule
		Base Model & \num{1.25e7} & \num{6.2E-02} \\ 
		KL Loss & --- & \num{4.7E-02} \\ \midrule
		Pyramid & \num{1.27e7} & \num{6.4E-02} \\ 
		Attention & \num{1.26e7} & \num{6.6E-02} \\
		Bidirectional & \num{1.35e7} & \num{6.7E-02} \\
		CNN & \num{1.66e7} & \num{5.8E-02} \\
		DMO-3 & \num{6.14e6} & \num{2.3E-01} \\
		\bottomrule
	\end{tabular}
	\caption{Comparison of state-of-the-art methods. Each row denotes a base model with only the method applied (i.e., one model component). KL loss was applied to the base model. DMO-3 denotes \textit{decoder multi-output} predicting 3 timesteps at a time. Reconstruction error is used as error.}
	\label{table:ablation}
\end{table}

Different state-of-the-art methods provide performance benefits to varying degrees. For example, DMO-3 exhibited the largest improvement. Not only does the DMO-3 model have fewer parameters, it can train faster.
The convolutional network during the encoder stage also provides additional benefits, likely due to the learned $7 \times 7$ convolutional kernels to better capture temporal context and reduce the number of timesteps.

\begin{figure}[t]
	\begin{center}
		\includegraphics[width=0.95\linewidth]{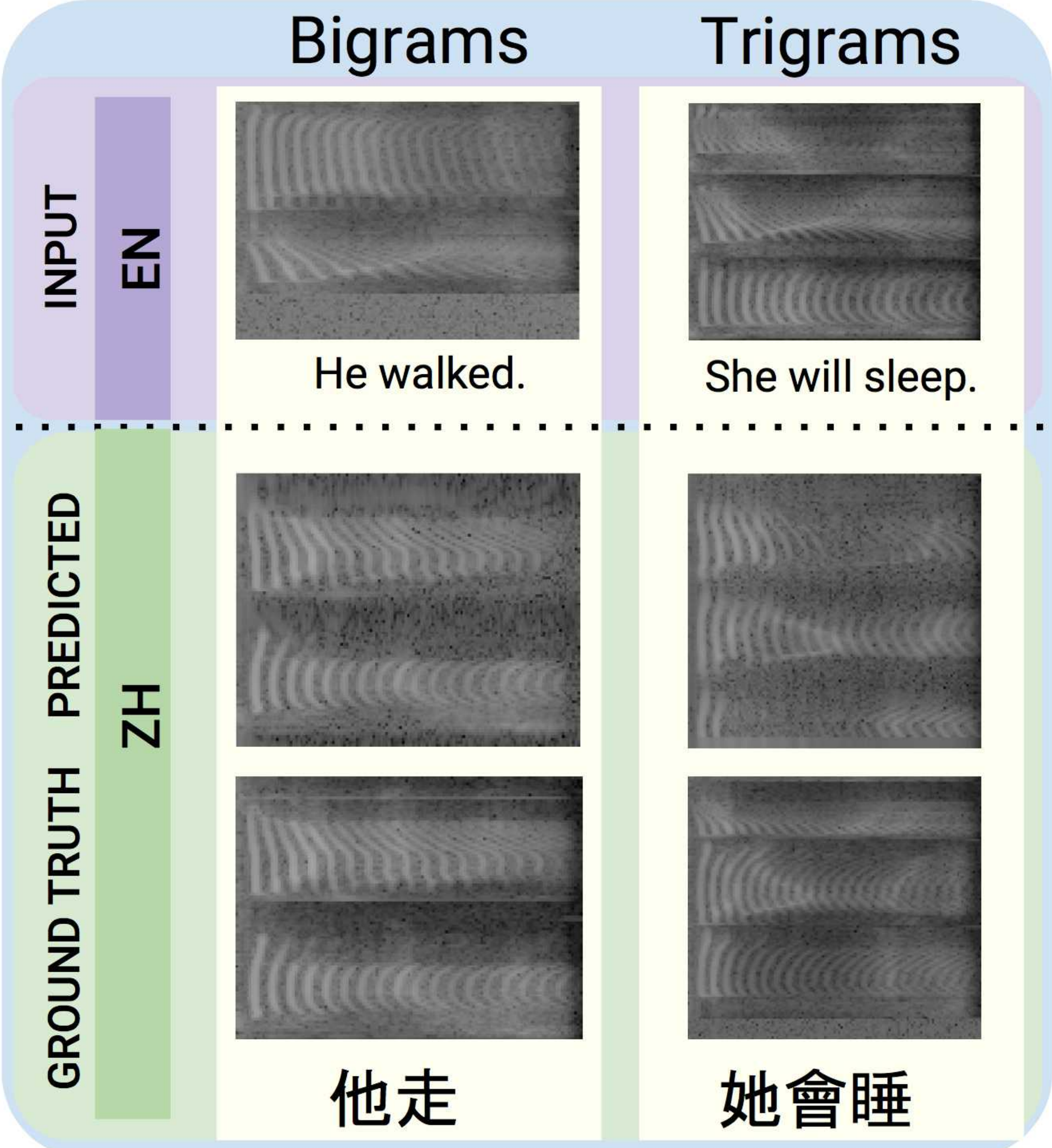}
	\end{center}
	\caption{Results on an unseen speaker for bigrams and trigrams. (Top row) Input English spectrogram and sentence. (Middle row) Predicted Mandarin spectrogram. (Bottom row) Ground truth Mandarin spectrogram and sentence.}
	\label{fig:ngrams}
\end{figure}

\subsection{Qualitative Results}

Figure \ref{fig:ngrams} shows the input, predicted, and ground truth spectrogram for both bigram and trigram inputs for our full model equipped with all components in Table \ref{table:ablation}.
The input spectrogram is a speaker not present in the training set. Figure \ref{fig:ngrams} represents our toughest experiment for our model; but the results are positive.
Our model is able to successfully generate rib-like patterns in the spectrogram.
Even for the trigram case, we are able to see three distinct words, delimited by silence.

\subsection{Learned Word Embeddings}

\begin{figure}[h]
\begin{center}
\includegraphics[width=0.7\linewidth]{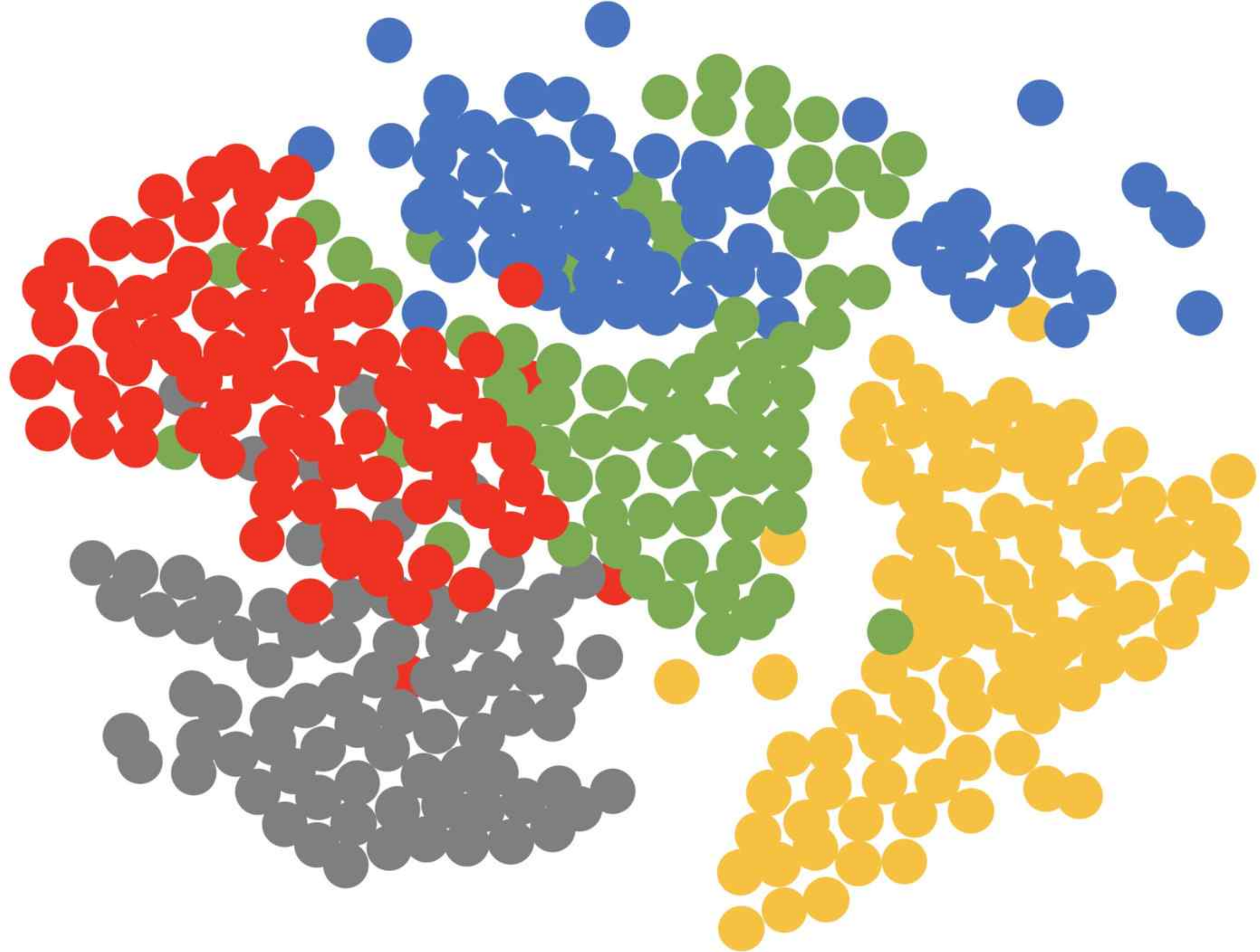}
\end{center}
   \caption{Learned word-level embeddings. Each dot is the encoder output state’s t-SNE 2D embedding. Colors denote different Spanish words.}
\label{fig:tsne}
\end{figure}

Internally, we trained our model on a dataset consisting of bigrams before moving to larger trigram-based models. The input was comprised of Spanish (ES) bigrams and the output consisted of English (EN) bigrams. The English vocabulary consisted of 70 words. We took all words in our input Spanish vocabulary and analyzed the convolutional activations in Figure \ref{fig:tsne}.
Each point represents a different instance of a word.

It is clear that our model can learn representations of individual Spanish words despite being trained on input biagrams in Spanish with output labels in English.

\section{Conclusion}
In computer vision, the community first made significant deep learning advances with on ImageNet \cite{russakovsky2015imagenet} by predicting single label classes \cite{krizhevsky2012imagenet}.
The community progressively moved to sequence outputs such as image captioning \cite{xu2015show, donahue2015long, karpathy2015deep}. Eventually, end-to-end models began to produce entire paragraphs of text describing a single image \cite{krause2016hierarchical}.

Similar to the story in computer vision, in this work, we presented a method for end-to-end spoken language translation on short phrases consisting of a few words.
Using a newly collected dataset of multiple speakers in multiple languages, our method is able to learn acoustic and language features while being able to generalize to unseen speakers.
We hope the speech and signal processing community will build on our work, moving to larger and more complex models for even longer sentences and full paragraphs.

\bibliography{main}
\bibliographystyle{acl_natbib}

\end{document}